# Target Pose Guided Whole-body Grasping Motion Generation for Digital Humans

Quanquan Shao and Yi Fang

*Abstract*— Grasping manipulation is a fundamental mode for human interaction with daily life objects. The synthesis of grasping motion is also greatly demanded in many applications such as animation and robotics. In objects grasping research field, most works focus on generating the last static grasping pose with a parallel gripper or dexterous hand. Grasping motion generation for the full arm especially for the full humanlike intelligent agent is still under-explored. In this work, we propose a grasping motion generation framework for digital human which is an anthropomorphic intelligent agent with high degrees of freedom in virtual world. Given an object known initial pose in 3D space, we first generate a target pose for whole-body digital human based on off-the-shelf target grasping pose generation methods. With an initial pose and this generated target pose, a transformer-based neural network is used to generate the whole grasping trajectory, which connects initial pose and target pose smoothly and naturally. Additionally, two post optimization components are designed to mitigates foot-skating issue and hand-object interpenetration separately. Experiments are conducted on GRAB dataset to demonstrate effectiveness of this proposed method for whole-body grasping motion generation with randomly placed unknown objects.

## I. INTRODUCTION

Objects grasping is a fundamental mode of humans to finish more complicated manipulation tasks in daily life. For various autonomous intelligent agents, it is vital to have the ability to grasp different types of objects with their manipulators or hands. The synthesis of grasping motion is widely demanded in many applications such as 3D games, virtual reality and robotics. In this work, the virtual digital human is selected as the anthropomorphic intelligent agent and attempt to endow them abilities to approach and grasp surrounding objects in virtual 3D space just like humans in real world. Because of the complexity of dexterous grasping manipulation, most works still focus on the generation of the static grasping pose with robotic dexterous multi-finger hand or humanoid hand while the approaching motion generation for object grasping is still under-explored especially for humanoid intelligent agents with high degrees of freedom. As a counterpart of humans in virtual world, digital humans not only need to be looked realistic but also need to act like humans including moving and interacting with objects.

*This research is supported by the Natural Science Foundation of China (Grant No. 52105031) and the Postdoctoral Research Foundation of China (Grant No. 2022T150407).

Quanquan Shao is with Hangzhou Huawei Enterprises Telecommunication Technologies Co., Ltd, Hangzhou 310056, China.

Yi Fang is with the Department of Automation, Shanghai Jiao Tong University, Shanghai 200240, China. (e-mail: apocalypse@sjtu.edu.cn).

Many researches have been done on generating human motion realistically based on text, audio or scenes. However, these works mostly focus on bodies without considering dexterous hands. It is a challenging task to generate full-body human motion including bodies and hands for approaching and grasping a target object in virtual 3D space. Firstly, integrating bodies and hands make the controlled virtual human have a quite higher action space. Following other pioneering works, we model the controlled human by SMPL-X[35]. Only considering bodies and hands, this model has 163 degrees of freedom for local pose with 51 spherical joints and 6 degrees of freedom for global translation and orientation. Secondly, the bodies and hands have different motion scales and the kinematic chain is rather long from foot joints contacting the floor to finger joints which may contacting objects. Thirdly, motion of the hand and fingers are strongly constrained by the shape of various objects to be grasped. Throughout the entire grasping motion, the hand should not penetrate into the target object and contact the object surface at last.

To deal with this challenging task, some solutions have been proposed by pioneer researchers. GOAL[36] tried to generate whole-body grasping motion by an auto-regressive fashion while it could not converge to the target pose precisely. The interpolation process of the last few frames caused heavy interpenetration and made the entire human motion unrealistic. SAGA[37] employed a marker-based representation of human pose and generated a grasping motion by a CNN network. However, it outputted a fixed-length motion trajectory no matter where the target object was placed in 3D space. On count of the marker-based representation and the constraint of object's shape, a time-consuming post-processing was also needed to obtain SMPL-X parameters. Here, we address this problem with a transformer-based motion generation network and two independent post-processing submodules. Given an object known initial pose in 3D space, we follow the target pose generation method of GOAL and only focus on the whole-body grasping motion synthesis problem. Conditional on a given initial pose and this generated target pose, the transformer encoder-based network could generate the whole grasping motion, which connects initial pose and target pose smoothly and naturally. An efficient inverse kinematics-based (IK-based) post-processing is proposed to reduce foot-skating. Benefited from the differentiability of SMPL-X, we could utilize a simple gradient descent algorithm to mitigate interpenetration of hand and objects to be grasped.

Contributions of this work are summarized as follows: 1) A well-behaved grasping motion generation framework for digital humans, which has a better performance and efficiency compared to previous works. 2) A time-efficient IK-based post-processing alleviating foot sliding when it contacts the ground. 3) A gradient descent-based optimization module for mitigating hand-object interpenetration while keeping the grasping motion smooth and the hand contacting the surface of the object at last frame of the sequence.

## II. RELATED WORK

### A. Dexterous Grasping Generation

Researches on dexterous grasping are mainly categorized into two types. One is analytical grasping method that models and optimizes the hand-object interaction in force-balance and mostly focuses on success rate of generated grasping pose[1,2]. These methods are normally time-consuming for dexterous grasping because of the high dimension of the search space. At the same time, outputted successful grasping poses are not limited to human-style. The other type is data-driven grasping outputting results strongly restricted by the training data[3,4]. Learning-based method is efficient but generated poses are not always reliable. Dexterous hands are diverse in robotics including Barrett Hand[5], Allegro Hand[6], ADROIT Hand[7] Shadow Hand[8]and so on, which have three to five fingers generally. A non-rigid human hand mesh model named MANO[9] is also used in dexterous grasping pose generation. As a parameterized model in computer graphics, it is more flexible and realistic to model various human hand pose and human grasping types[10]. Compared to directly outputting grasping poses[3], intermediate representations of grasping poses such as contact points[11,12] and contact maps[13] are also applied for grasping generation models. Then some optimizations are performed to obtain grasping poses with these intermediate representations.

For grasping manipulation, it is not enough to generate a static target grasping pose. Grasping motions from initial pose to target pose even subsequent manipulation trajectories are also need to be taken into consideration. Rajeswaran et al. used deep reinforcement learning (DRL) to learn grasping skill and other more complicated manipulation skills with human-like ADROIT hand in simulation[14].GRAFF combined image affordance map with reinforcement learning for generating grasping motion with a dexterous robotic hand considering human preference[15]. DRL-based grasping manipulation skills are object shape-specific and may grasp objects in a non-human style. DexMV[16] utilized GAIL[17] framework for robotic manipulation skill learning. It obtained object-hand interaction trajectories from human demonstration videos and retargeted these sequences to a robotic dexterous hand. Robots could obtain more human-like skills based on this framework. D-grasp generated a whole grasping motion guided by a coarse target pose in a simulator[18]. The motion generation problem of two-hands collaborative manipulation was also explored in ManipNet[19]. This method outputted trajectories of each hand modeled by MANO when the motion sequences of the target object and two hand wrists were given. Fine-tuning approaches for coarse manipulation motions are also explored based on hand-object distance[20] or physical simulator[21].

### B. Human Motion Generation

Methods of human motion Generation aim to generate natural, realistic and various 3D human motions based on different conditional information[22]. Action2Motion[23] generated diverse human motions with a VAE-based network conditioned on different action labels that were represented by one-hot vectors. Tevet et al. proposed a framework named Motionclip for generating human motions following natural language instructions[24]. Music-conditioned human motion generation namely dancing to the music was also explored via a full-attention cross-modal transformer-based network [25]. GAMMA which was a goal position-guided human motion generation framework could be extended to infinite long human locomotion easily[26]. However, human motion was all generated in free 3D space based on these methods.

Human motion generation for interaction between human and their surroundings is also an important aspect. Wang et al. embedded scene points into feature vectors and inputted them as conditions into human motion generative model in 3D space[27]. SAMP[28] first generated the target position and then integrated path planning methods with learning-based motion generation network to generate motion of siting on the chair, lying on the sofa, etc. Mir et al.[29] proposed a motion synthesis method considering 3D scene constraint, which first generated a root path trajectory with motion planning method and then output a whole motion sequence following the root path. A goal-centric canonical coordinate frame was used for motion generation and could generate long motion sequences of diverse actions, which not affected by global positions. To promote researches of the scene-conditioned human motion generation, serval datasets about human motion generation in 3D scene were also released including PROX[30], SAMP[28], COUCH[31], HUMANISE [32], CIRCLE[33]and so on. These datasets and methods chose SMPL[34]or SMPL-X[35] without hand pose as the human model and mainly focused on the interaction of bodies and the scene.

Taking poses of human hand and object-hand interaction into account, GRAB dataset released more than one thousand whole-body grasping and manipulation trajectories[3]. Some works have been done based on this dataset[36-38]. Similar to our task, GOAL[36] and SAGA[37] focused on grasping motion generation. GOAL utilized an auto-regressive network and could not converge to the target pose precisely. SAGA only generated fixed-length whole-body grasping motion sequence with a CNN network. IMOS[38] was designed for generating the entire manipulation trajectories. It mainly focused on motion generation but had heavy artifacts of the interaction of objects and the hand in generated sequence.

## III. METHODOLOGY

In this section, we will introduce the proposed method in detail. The goal of this task is generating natural, realistic human grasping motion conditioned on the initial human pose, the target human pose and the object to be grasped. It mainly includes three modules, a transformer-based generation network, an IK-based lower-body motion post-processing module relieving foot-skating and a gradient descent-based upper-body grasping motion post-processing to alleviate the interpenetration of the human hand and objects. SMPL-X is used as the human model in this work, which models full body mesh with hand pose. Ignoring expression parameters, the SMPL-X parameters include the shape parameters $\beta \in R^{10}$, the full-body local pose parameters $\theta \in R^{51 \times 3}$, the global transition parameters $t \in R^3$ and the global orientation parameters $\varphi \in R^3$. The whole body pose parameters are further divided into the body pose $\theta_b \in R^{21 \times 3}$, the right hand pose $\theta_r \in R^{15 \times 3}$ and the left hand pose $\theta_l \in R^{15 \times 3}$. For this grasping motion generation task, we only consider grasping objects with right hand and spherical joints are represented by continuous 6D vectors[39]. For a given character, the full-body pose can be represented as $X = [t, \varphi, \theta_b, \theta_r] \in R^{225}$. The grasping motion sequence can be symbolized as $\Gamma = [X_0 ... X_T]$. T represents the length of this motion sequence and is a hyperparameter in this situation.

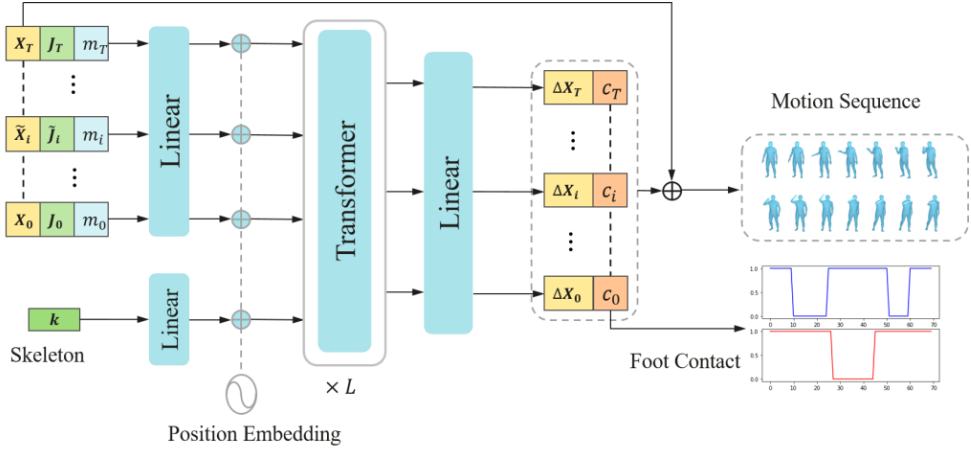

Fig. 1: The transformer-based motion generation framework

### A. Transformer-based Motion Generation

The purpose of this module is to generate a sequence of human poses $\Gamma$ given an initial pose $X_0$ and the target pose $X_T$ of a specific virtual human character. Detail structure of this transformer-based network is described in Fig. 1. We choose bone length vectors of this human $k \in R^{41}$ identifying the specified character rather than the original shape parameters $\beta$ on count of length of bones is more relevant to kinematic motion sequence. Skeleton parameters $k$ are composed of length of bones of the body and the right hand in SMPL-X. In Addition, the human pose $X$ is extended with global positions of these joints $J \in R^{37 \times 3}$, written as $Z = [X, J]$. With the initial pose $Z_0$ and the target pose $Z_T$, we generate a naive initial motion sequence $Z_0, ..., \tilde{Z}_i, ..., Z_T$ using a simple linearly interpolation from $Z_0$ to $Z_T$. And then an extra flag element $m_i$ is added to each pose to distinguish the given pose and the interpolated pose. It is set to zero and one respectively. Sequence of this extended human pose $[Z_i, m_i]$ and the skeleton $k$ are inputted into the transformer-based motion generation network. Outputs of this network are probabilities of two foot contacting the ground and the deviation $\Delta X$ of each human pose. For training this network some losses are applied. The first loss is 1-norm or 2-norm of output terms of the network, which is defined as,

$$L_1 = \omega_t \|t_i - \hat{t}_i\|_1 + \omega_{rg}\|\varphi_i - \hat{\varphi}_i\|_1 + \omega_{rb}\|\theta_{b_i} - \widehat{\theta_{b_i}}\|_1 + \omega_{rh}\|\theta_{r_i} - \widehat{\theta_{r_i}}\|_1 + \omega_{fc}\|c_i - \hat{c}_i\|_2 \qquad (1)$$

where $\hat{t}_i, \hat{\varphi}_i, \widehat{\theta_{b_i}}, \widehat{\theta_{r_i}}, \hat{c}_i$ are predicted translations, global orientations, body poses, hand poses and the probabilities of foot contacting the ground. $\omega_t, \omega_{rg}, \omega_{rb}, \omega_{rh}$ and $\omega_{fc}$ are weights of each term respectively. To improve smoothness of the generated motion sequence, the loss of the deviation of rotate angles represented by 6D vectors is also added, which is formulated as,

$$L_2 = \|\Delta\varphi_i - \Delta\hat{\varphi}_i\|_1 + \|\Delta\theta_{b_i} - \Delta\widehat{\theta_{b_i}}\|_1 + \|\Delta\theta_{r_i} - \Delta\widehat{\theta_{r_i}}\|_1,$$

$$\Delta\gamma_i = \gamma_i - \gamma_{i-1}, \gamma = \{\varphi, \theta_b, \theta_r\}. \qquad (2)$$

Taking advantage of the differentiability of SMPL-X, we also consider the loss of joints defined as,

$$L_3 = \|J_i - \hat{J}_i\|_1 + \|\Delta J_i - \Delta\hat{J}_i\|_1 + \|\Delta^2 J_i - \Delta^2\hat{J}_i\|_1 \qquad (3)$$

$$\hat{J}_i = smplx(\hat{t}_i, \hat{\varphi}_i, \widehat{\theta_{b_i}}, \widehat{\theta_{r_i}}) \qquad (4)$$

$$\Delta J_i = J_i - J_{i-1}, \Delta^2 J_i = \Delta J_i - \Delta J_{i-1} \qquad (5)$$

where $smplx$ means the differentiable SMPL-X model. $\hat{J}_i, J_i$ are global positions of joints of predicted human pose and the ground truth of joint positions respectively. The loss of positions of finger joints in hand local coordinate system is applied for paying more attention to motion of fingers in grasping sequence, which is written as,

$$L_4 = \|J_{i,h} - \hat{J}_{i,h}\|_1 \qquad (6)$$

where $J_{i,h}$ and $\hat{J}_{i,h}$ are positions of finger joints in local wrist frame. At last, the total training loss of this model is,

$$L_{total} = L_1 + \omega_2 L_2 + \omega_3 L_3 + \omega_4 L_4 \qquad (7)$$

where $\omega_2, \omega_3, \omega_4$ represent the loss weights of each term.

### B. IK-based Lower-body Motion Post-processing

The goal of this module is to alleviate foot-skating of the generated human motion sequence. The skeleton of SMPL-X model is shown in Fig. 2(d). We fix joints above the root joint and only adjust two legs separately. The bone chain of one leg is illustrated in Fig. 2(c), which includes hip joint, knee joint, ankle joint and foot joint. Inverse kinematics(IK) approach is widely applied in robotics and animation. Many popular IK methods have been proposed such as CCD[40]. In this work, we choose the two-bone-IK[41] method that has two bones and three joints. With this simple structure, it has an analytical solution obtained with basic knowledge of geometry.

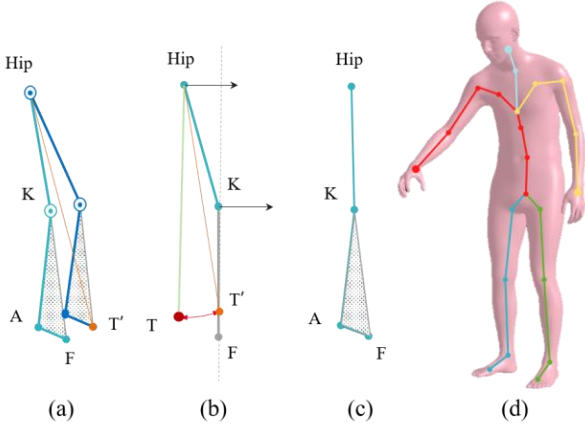

Fig. 2: Diagrams of IK-based lower-body motion post-processing

This IK-based module has three steps. Firstly, based on a threshold, the predicted contacting probabilities of two foots are changed into Boolean value, shown in Fig. 1. Foot joints of the generated pose sequence are grouped according to whether they are in continuous contact with the ground. The foot joint position of the start pose is selected as the target of the subgroup including the start pose, and we do the same for the target pose. For subgroups with middle frames, the mean position of foot joints in each subgroup is regarded as the new position of these foot joints.

Secondly, since one leg has three links and four joints, we fix the ankle joint to convert it to the normal structure of two-bone-IK, shown in Fig. 2(c). The target of this step is to translate the foot joint to a new target position $\mathbf{T}$ for each pose, illustrated in Fig. 2. Different from traditional two-bone-IK, K-A-F plane is selected as the rotation plane and this operation could obtain a smoother refined motion in this situation. The hip joint is projected into K-A-F plane while target position $\mathbf{T}$ is rotated about an axis perpendicular to the plane of the paper to obtain an intermediate target position $\mathbf{T}'$ in K-A-F plane, which is depicted in Fig. 2(b). A normal two-bone-IK method is applied for Hip-K-F kinematic chain in K-A-F plane and moves the foot joint from F to $\mathbf{T}'$ in Fig. 2(a). Another rotation transformation is followed to rotate $\mathbf{T}'$ to $\mathbf{T}$.

Thirdly, these subsequences that do not contact the ground also need to be changed accordingly for keeping the overall motion sequence smooth. A subsequence of one joint can be written as $[\boldsymbol{\theta}_2, \ldots, \boldsymbol{\theta}_{K-1}]$. It is extended with two angles of adjacent pose contacting the ground, which can be represented as $[\boldsymbol{\theta}_1, \boldsymbol{\theta}_2, \ldots, \boldsymbol{\theta}_{K-1}, \boldsymbol{\theta}_K]$. A mean filter of size 3 is applied for this subsequence at first. Then with the refined $\boldsymbol{\theta}_{n,1}$ and $\boldsymbol{\theta}_{n,K}$ based on previous two-bone-IK, a new subsequence of the joint angle can be obtained following

$$\boldsymbol{\theta}'_{old,i} = \boldsymbol{\theta}_{old,i} + (\boldsymbol{\theta}_{n,1} - \boldsymbol{\theta}_{old,1}) \quad (8)$$

$$\boldsymbol{\theta}_{n,i} = \boldsymbol{\theta}'_{old,i} + \frac{i-1}{K-1}(\boldsymbol{\theta}_{n,K} - \boldsymbol{\theta}'_{old,K}), i = 1, \ldots, K \quad (9)$$

where $\boldsymbol{\theta}_{old,i}$ is joint angle generated by the transformer-based network and $\boldsymbol{\theta}_{n,i}$ is refined joint angle. It should be noticed that only hip joint and knee joint are changed in this IK-based lower-body processing.

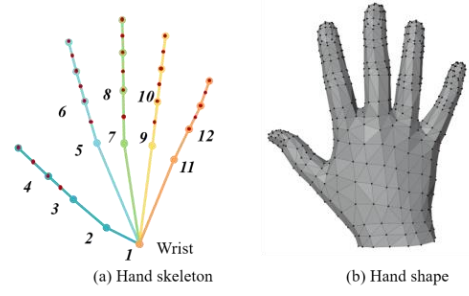

Fig. 3: The hand skeleton and shape model.

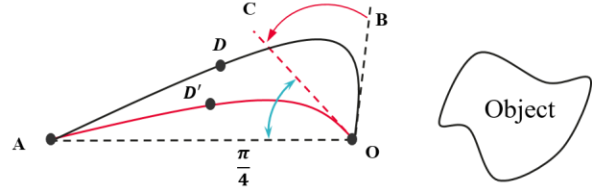

Fig. 4: Improvement of hand wrist trajectory.

### C. Upper-body Grasping Motion Post-processing

To mitigate interpenetration of the target object and the hand of SMPL-X, some additional processing is required. The hand mesh of SMPL-X, namely MANO, has 778 vertices and 1554 triangular faces, shown in Fig. 3. Making use of differentiability of SMPL-X, we apply a simple gradient descent optimization to refine the distance between vertices of hand and point cloud of the target object. It is harmful to smoothness of motion sequence to optimize the full parameters of SMPL-X $[\boldsymbol{t}, \boldsymbol{\varphi}, \boldsymbol{\theta}_b, \boldsymbol{\theta}_r]$. Instead, we change sequence of global position of wrist joint empirically and only optimize some hand-related joints that are numbered 1 to 12 in Fig. 3(a). As illustrated in Fig. 4, $\boldsymbol{O}$ is the target position of a wrist and $\boldsymbol{A}$ is a wrist of the first pose to be refined that is selected based on a threshold. In our experiment, wrists 0.4m close to the target position $\boldsymbol{O}$ are all corrected by this operation. If the maximum angle $\angle AOB$ of the generated wrist sequence is larger than $\frac{\pi}{4}$, we rotate each wrist $\boldsymbol{D}$ to $\boldsymbol{D}'$ with

$$\angle AOD' = \frac{\angle AOD}{\angle AOB} * \frac{\pi}{4} \quad (10)$$

The purpose of this operation is to constrain the generated wrist sequence within a cone with half-angle of $\frac{\pi}{4}$, which could reduce hand-object penetration in some degree. Based on these updated hand wrists, two-bone-IK method is applied to this shoulder-elbow-wrist chain.

The object is represented with point cloud of size 4096 in global coordination, marked as $\boldsymbol{P}_o \in \boldsymbol{R}^{4096 \times 3}$. A signed Chamfer Distance[42] between $\boldsymbol{P}_o$ and hand vertices $\boldsymbol{V}_h \in \boldsymbol{R}^{778 \times 3}$ is used as a collision loss,

$$E_1 = \alpha_1 \sum_{i=1}^{L} d\left(S^{\pm}(\boldsymbol{P}_{o,i}, \boldsymbol{V}_{h,i})\right) + \alpha_2 \sum_{i=1}^{L} d\left(S^{\pm}(\boldsymbol{V}_{h,i}, \boldsymbol{P}_{o,i})\right),$$
$$d(x) = abs(min(x + \delta, 0)) \quad (11)$$

where $S^{\pm}(\boldsymbol{P}_o, \boldsymbol{V}_h)$ is the signed Chamfer Distance from $\boldsymbol{P}_o$ to $\boldsymbol{V}_h$, $\delta$ is a tolerance threshold, $\alpha_1$ and $\alpha_2$ are weights of different terms. $L$ means number of the last poses that will be optimized. For the last frame of the generated grasping motion, the hand should also contact to the object surface. Each finger is considered independently and Chamfer Distance of vertices of one finger $V_{f,g}$ and $P_o$ is optimized to zero with a contact loss,

$$E_2 = \alpha_3 \sum_{g=1}^{N_f} S(V_{f,g}, P_{o,i}) \quad (12)$$

where $N_f$ is number of fingers to contact the object. We constrain 4 fingers without the pinky finger to contact the object in this situation. Avoiding interpenetration of different fingers, the distance of different fingers is also constrained. We choose the joints and the middle points of bones in each finger, which are red points shown in Fig. 3. Loss of the distance of different fingers is applied,

$$E_3 = \alpha_4 \sum_{g=1}^{5} abs(min(S(H_{f,g}, H_{f,\Omega}) - \delta_2, 0)) \quad (13)$$

where $H_{f,g}$ is points in finger $g$ and $H_{f,\Omega}$ are points of other four fingers, $\delta_2$ is a minimum distance of different fingers, which is set to 1.2cm here. Considering about the smoothness of grasping sequence, a deviation loss of finger joints is used,

$$E_4 = \alpha_5 \sum_{i=1}^{L-1} abs(min(v_{i,h} - \hat{v}_{i,h}, 0)), \quad (14)$$

$$v_{i,h} = \|J_{i,h} - J_{i-1,h}\|_2 \quad (15)$$

where to $v_{i,h}$ is the deviation of updated hand joints, $\hat{v}_{i,h}$ is deviation of hand joints of the generated motion sequence. With these losses, a few gradient descent operations are executed to refine penetration of hand and the object. At last a smooth mean filter of size 3 is applied for these changed angles of joints of the arm and the hand.

TABLE I. Comparison Of Different Methods

| Models | END-MJD(mm) | | PSKL-J | | INTER-VOLUME(cm³) | | | SKATING (cm/s) ↓ | TIME (s) ↓ |
|---|---|---|---|---|---|---|---|---|---|
| | *Body*↓ | *rHand*↓ | *Pred,GT*↓ | *GT,Pred*↓ | *V1*↓ | *V5*↓ | *V10*↓ | | |
| GOAL[36] | 75.82 | 32.02 | 0.6827 | 0.6334 | / | / | / | 2.06 | 4.73e+1 |
| SAGA[37] | 5.50 | 7.24 | 0.4900 | 0.5020 | 4.22 | 4.44 | 4.44 | 4.66 | 7.71e+2 |
| Ours w/o Opt. | 2.23 | 2.92 | 0.2762 | 0.2511 | 2.67 | 4.25 | 4.44 | 4.72 | 2.78e-2 |
| Ours+p1 | 2.22 | 2.94 | 0.2112 | 0.2578 | 2.72 | 3.88 | 4.07 | 2.77 | 2.93e-2 |
| Ours+p1+p2 | 7.04 | 2.94 | 0.2349 | 0.2568 | 2.69 | 3.87 | 4.05 | 1.68 | 8.89e-1 |
| Ours+p1+p2+p3 | 7.11 | 5.03 | 0.2418 | 0.2627 | 0.86 | 0.98 | 1.11 | 1.69 | 3.65e+1 |

## IV. Experiments

This transformer-based network is trained with GRAB[3] dataset, which is split into training dataset and test dataset following GOAL[36]. It has 8 layers with 512 dimensions with 4 headers and is trained totally for 100K iterations with Adam optimizer. The learning rate is 1e-4 and $\omega_t, \omega_{rg}, \omega_{rb}, \omega_{rh}, \omega_{fc}, \omega_2, \omega_3, \omega_4$ are set to 1,1,1,1,1,1,5,20, respectively. Parameters of post-processing $\alpha_1, \alpha_2, \alpha_3, \alpha_4, \alpha_5, L$ and $\delta$ are set to 1,1,50,100,10, 15 and 0.004. Forty iterations are updated in gradient descent optimization to alleviate interpenetration between the human hand and the object.

Some metrics are used for evaluate our proposed method. 1) **END-MJD**. It represents the mean joint distance between end pose of the generated motion sequence and the given target pose. This indicator measures the degree of controllability of this proposed method. According to the joints of human body and the right hand, two subclasses are calculated separately. 2) **PKSL-J**. For motion smoothness and similarity between the generated motion and the training data, we use PKSL-J that means the Power Spectrum KL divergence between joint acceleration distribution of the synthesized motion and the test dataset, following [43]. As the asymmetry of KL divergence, results of both directions are applied. 3) **INTER-VOLUME**. It means interpenetration volumes between the right hand and the object. We voxelize the object and the right hand with grid step size of 0.5cm and the intersection volume can be obtained with these voxels. Maximum volume of the last frame, the last five frames and the last ten frames are all evaluated, i.e. *V1*, *V5* and *V10*. 4) **SKATING**. Foot-skating artifact is also a very important issue for motion generation. In this work, the minimum joint velocity of two foots is used as the index of foot-skating for a frame. The summation of all frames in a sequence measures degree of foot skating of the whole motion. 5) **TIME**. Time cost of a pipeline is another performance indicator. We evaluate the total time cost of different methods including network inference time, pre-processing time and post-processing time.

### A. Comparison

GOAL[36] and SAGA[37] are used as baselines. GOAL is deployed just like descriptions of the paper. For comparison, network of SAGA is retrained with GRAB dataset that is split into training and test dataset same as GOAL. Experimental results can be found in Tab. I. "Ours w/o Opt." means performance of the raw output of the transformer-base network. "Ours+p1" represents that the output is processed by a simple mean filter of size 3. "Ours+p1+p2" denotes that the output is additionally processed with the proposed IK-based lower-body post-processing. "Ours+p1+p2+p3" means that the gradient descent-based optimization is also applied for this generated motion sequence. For "END-MJD" indicator, our method has the best performance while mean distance of hand joints of GOAL is more than 30mm that indicates this approach cannot converge to the target pose successfully. For the smoothness evaluation of generated motion, the proposed method has significant advantage with about 0.25 in PSKL-J compared to GOAL and SAGA that are all larger than 0.5. Considering the index of interpenetration of the hand mesh and the object mesh, we can find that this gradient descent-based optimization method can effectively alleviate the interpenetration without significantly worsening other evaluation indicators. SAGA has the worst performance for the foot-skating issue while our proposed IK-based lower-body motion post-processing could reduce the indicator of foot-skating from 2.77cm/s to 1.67cm/s, which is quite effective. As for the evaluation of time efficiency, SAGA is extremely time-consuming, which has a heavy post-processing. We can find that our proposed IK-based lower-body post-progressing is time-efficient with less than 0.9s while the gradient descent-based upper-body post-progressing is a bit time-consuming, taking about 36 seconds for 40 iterations. Overall, compared to baselines, the proposed method still has the best time performance including all post-progressing. Qualitative results are illustrated in Fig. 5.

TABLE II. RESULT OF DIFFERENT LOSS FUNCTION

| Models | END-MJD(mm) | | PSKL-J | | INTER-VOLUME(cm³) | | | SKATING (cm/s)↓ |
|---|---|---|---|---|---|---|---|---|
| | Body↓ | rHand↓ | Pred,GT↓ | GT,Pred↓ | V1↓ | V5↓ | V10↓ | |
| Ours | 2.23 | 2.92 | 0.2762 | 0.2511 | 2.67 | 4.25 | 4.44 | 4.72 |
| Ours w/o. $L_4$ | 3.25 | 3.18 | 0.2541 | 0.2195 | 2.67 | 4.39 | 4.69 | 4.72 |
| Ours w/o. $L_3, L_4$ | 2.98 | 3.92 | 0.4759 | 0.4685 | 2.76 | 4.93 | 5.51 | 6.06 |
| Ours w/o. $L_2, L_3, L_4$ | 3.72 | 6.61 | 0.7068 | 0.7326 | 3.53 | 5.37 | 5.80 | 5.55 |

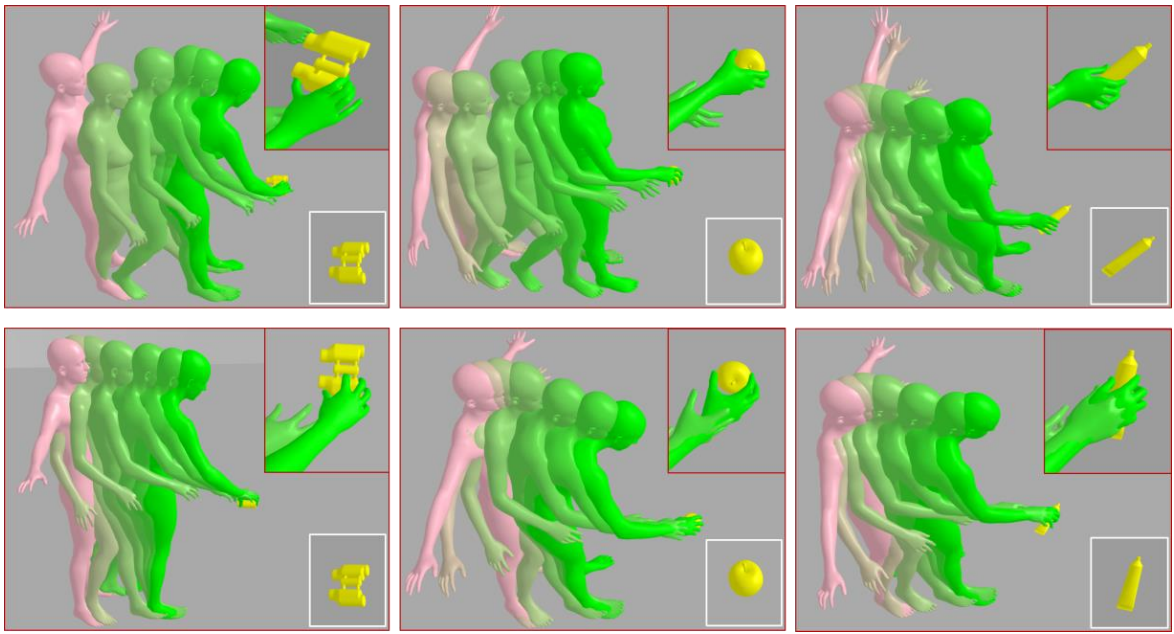

Fig. 5: Generated grasping motion for different objects.

## B. Ablation Study

For the motion generation network, there are four terms of the training loss. We conduct several ablation experiments to study the effects of these term. Models with the same structure are trained based on combination of different loss. Evaluation metrics are selected same as comparative studies except the indicator of time cost and detailed results are shown in Tab. II. Comparing "Ours" with "Ours w/o $L_4$", we find the loss of positions of finger joints in hand local frame, namely $L_4$, can improve the control precision of hand joints slightly and reduce interpenetration of the hand and the object in grasping motion in some degree. With the differentiability of SMPL-X, we find that the joint position-related loss $L_3$ can enhance the smoothness of generated motion and also mitigate the foot-skating issue at the same time. The loss of the deviation of rotate angles represented by 6D vectors i.e. $L_2$ remarkably improves the smoothness of generated human motion while may worsen the foot-skating problem.

## V. CONCLUSION

In this work, a grasping motion generation framework for digital human in virtual world is proposed. A transformer-based network is applied to generate grasping human motion sequence. Two post optimization components are designed to alleviate foot-skating and hand-object interpenetration.